\documentclass[conference]{IEEEtran}
\IEEEoverridecommandlockouts
\usepackage{lastpage}
\usepackage{fancyhdr}
\usepackage{setspace} 
\usepackage{array}
\usepackage{caption}
\usepackage{cite}
\usepackage{amsmath,amssymb,amsfonts}
\usepackage{graphicx}
\usepackage{textcomp}
\usepackage[table,xcdraw]{xcolor}
\usepackage{listings}
\usepackage{lipsum}
\usepackage{verbatimbox}
\usepackage[euler]{textgreek}
\usepackage{gensymb}
\usepackage{multirow}

\usepackage{dsfont}

\usepackage{mathrsfs}
\usepackage{algorithmic}
\usepackage{graphicx}
\usepackage{textcomp}
\usepackage{ragged2e}
\usepackage{bbm}
\usepackage{diagbox}
\usepackage[linesnumbered, ruled]{algorithm2e}
\usepackage[caption=false, font=footnotesize]{subfig}
\usepackage{mathtools} 
\usepackage[hang,flushmargin]{footmisc}
\usepackage{lipsum}
\def\BibTeX{{\rm B\kern-.05em{\sc i\kern-.025em b}\kern-.08em
    T\kern-.1667em\lower.7ex\hbox{E}\kern-.125emX}}
\usepackage{hyperref}
\hypersetup{
    colorlinks=true,
    linkcolor=magenta,
    filecolor=blue,      
    urlcolor=cyan,
    citecolor=green,
}

\begin{document}

\title{ An Optimized YOLOv5 Based Approach For Real-time Vehicle Detection At Road Intersections Using Fisheye Cameras}

\author{Md. Jahin Alam,
        Muhammad Zubair Hasan,
        Md Maisoon Rahman,
        Md Awsafur Rahman,
        Najibul Haque Sarker,\\
        Shariar Azad,
        Tasnim Nishat Islam,
        Bishmoy Paul,
        Tanvir Anjum,
        Barproda Halder,
        Shaikh Anowarul Fattah
}



%


\maketitle

\begin{abstract}
Real time vehicle detection is a challenging task for urban traffic surveillance. Increase in urbanization leads to increase in accidents and traffic congestion in junction areas resulting in delayed travel time. In order to solve these problems, an intelligent system utilizing automatic detection and tracking system is significant. But this becomes a challenging task at road intersection areas which require a wide range of field view. For this reason, fish eye cameras are widely used in real time vehicle detection purpose to provide large area coverage and 360 degree view at junctions. However, it introduces challenges such as light glare from vehicles and street lights, shadow, non-linear distortion, scaling issues of vehicles and proper localization of small vehicles. To overcome each of these challenges, a modified YOLOv5 object detection scheme is proposed. YOLOv5 is a deep learning oriented convolutional neural network (CNN) based object detection method. The proposed scheme for detecting vehicles in fish-eye images consists of a light-weight day-night CNN classifier so that two different solutions can be implemented to address the day-night detection issues. Furthurmore, challenging instances are upsampled in the dataset for proper localization of vehicles and later on the detection model is ensembled and trained in different combination of vehicle datasets for better generalization, detection and accuracy. For testing, a real world fisheye dataset provided by the Video and Image Processing (VIP) Cup organizer ISSD has been used which includes images from video clips of different fisheye cameras at junction of different cities during day and night time. Experimental results show that our proposed model has outperformed the YOLOv5 model on the dataset by 13.7\% mAP @ 0.5.
\end{abstract}
\IEEEoverridecommandlockouts

%
\IEEEpeerreviewmaketitle

\section{Introduction}
Transportation systems are currently an indispensable part of human activities. It is estimated that on an average forty percent of the population spends at least an hour on the road each day \cite{vehicle-stat}. Vehicle detection and tracking for traffic surveillance is an active research topic in computer vision \cite{hu2004traffic} for solving traffic problems like, emergency braking \cite{milanes2012vision}, road congestion, wrong lane occupancy through accessing road-traffic intensity, automated route planning and number of on-road vehicles etc.
A pre-requisite for enabling this analysis is to accurately locate vehicles in video images so that vehicle attributes can be extracted, compared, and then counted. However, the efficiency and accuracy of vehicle analysis are seriously affected by complexities of vehicle appearances and traffic scenarios.

Contemporary vehicle detection methods, as proposed in literature reports, include background/foreground subtraction, frame differencing \cite{ego-estimation}, hand-crafted feature based methods \cite{motion}, neural-network based approaches (CNN)\cite{xuze2018cnn} and reinforcement learning\cite{zhang2020using}. 
Due to abundance of labelled data set with sufficient varieties  and considerable state-of-the-art performance of neural network based approaches, the convolutional neural network(CNN) based architectures are most commonly used as the vehicle detection method\cite{bochkovskiy2020yolov4}. Conventional CNN based object detection methods include 
R-CNN \cite{girshick2014rich},
Fast-R-CNN \cite{girshick2015fast},
Faster R-CNN \cite{ren2015faster},
SSD \cite{liu2016ssd},
DSSD  \cite{fu2017dssd},
R-FCN \cite{dai2016r},
FPN FRCN \cite{lin2017feature},
RetinaNet \cite{lin2017focal},
YOLOv3 \cite{DBLP:journals/corr/abs-1804-02767}, 
YOLOv4 \cite{bochkovskiy2020yolov4},
Efficientdet \cite{tan2020efficientdet}.
For fast as well as accurate object detection in real-time, various YOLO  models are considered to be the most suitable  \cite{s19030594} \cite{bochkovskiy2020yolov4}. YOLOv3 \cite{DBLP:journals/corr/abs-1804-02767} model is surpassed in both speed and accuracy by its successor YOLOv4 \cite{bochkovskiy2020yolov4} model. Later on YOLOv5 \cite{repo} is published with slight modifications from the YOLOv4 architecture.



The  Convolutional Neural Networks (CNN) based methods generally require large annotated datasets to make them scene agnostic. Moreover it is difficult to ensure that these methods detect the  objects purely based on motion cues and not overfit to appearance cues of commonly occurring moving objects, like vehicles or pedestrians\cite{CNN-large}.


Automated surveillance systems demand large field of view, to cover a wider area. A large field of view is generally attained by using various types of cameras such as, synchronized multiple cameras \cite{koutsia2008intelligent}, pan till zoom(PTZ) camera and fish eye camera. Among them, fish eye camera provides significant improvement in field of view (generally 180 degrees to 360 degrees), by just replacing the camera lens with fish eye lense. Because of having 360 degree field of view, it leaves no blind spot in sight and the overall surveillance system expense is alleviated with respect to the alternatives, as it can substitute multiple cameras all at once \cite{kim2016fisheye}\cite{wang2015real}. Additionally, being able to withstand rough weather, having night vision and water resistance make these particular type of cameras ideal for surveillance at road intersection areas. 


\begin{figure*}[t]
    \centering
    \includegraphics[scale=0.27]{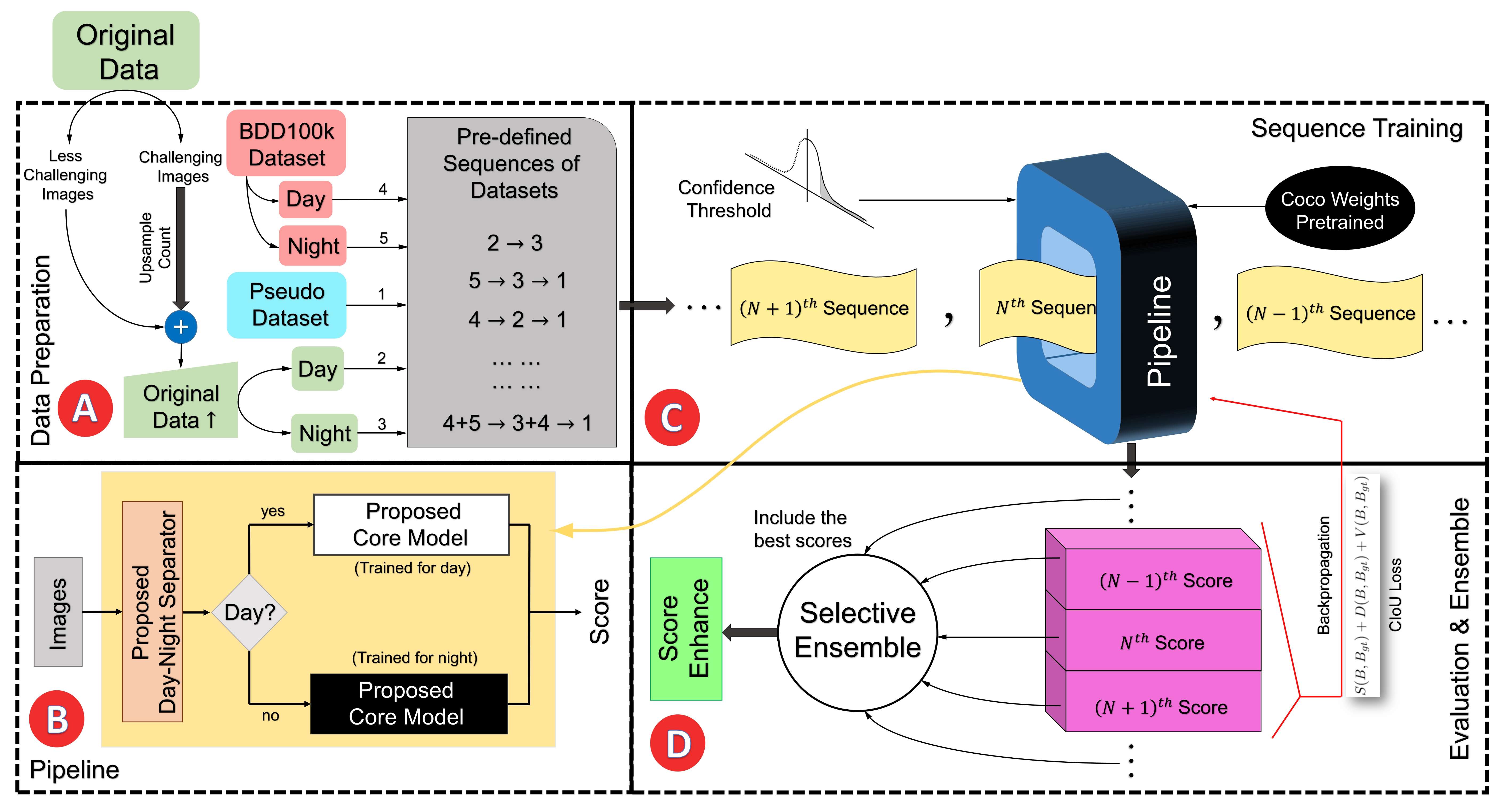}
    \caption{\textbf{Graphical Abstract of the proposed scheme: The Challenging subset of images are up-sampled in number and combined with the rest before creating \textcolor{blue}{data partitions} (1, 2, 3, 4, 5). From this, \textcolor{blue}{sequential partitions} are defined and each sequence is passed onto the implemented \textcolor{blue}{Pipeline} which generalizes the weights of the core model. \textcolor{blue}{Scores or evaluation metrics} produced by each sequence are looked over carefully and the best ones are selected for ensemble.}}
    \label{GA}
\end{figure*}

Utilizing fish-eye camera imposes some unique challenges. The fish eye distortion and environmental effects lead to deteriorated performance in detecting vehicles that are away from the camera or shaded by obstacles, such as trees and traffic light stands. In night time, surrounding lighting conditions and additional vehicle light exposures blur the image, causing further degraded detection accuracy. Conventionally, to compensate for the distortions, fisheye image based vehicle detection is done by first un-warping the image and then using a classifier \cite{ bertozzi2015360}, estimation of state parameters\cite{ tadjine2011vehicle}, correct fisheye distortion using cylindrical model \cite{ dooley2015blind }. In the existing approaches of object detection in fish-eye images,  managing barrel distortions is difficult \cite{goodarzi2019optimization}. As the vehicle moves away from the pinhole-center of the image, the distortion strength increases, causing the shape of the vehicle to change \cite{bertozzi2015360}. Normal digital image processing fails to achieve ideal results for fish-eye image processing, due to these non-linear distortions.

In this paper, a methodology is presented in order to detect road-intersection vehicles in fisheye images. This scheme presents a solution in managing highly accurate detetion from a pool of miscellaneous varieties of vehicles. The major contributions can be summarized as such:
\begin{enumerate}
    \item A new fisheye data set published in IEEE VIP Cup 2020 has been used that contains both day time and night time images, collected at different junctions with different environment and installation conditions.
    
    \item A day-night image separator has been implemented to overcome the distortion introduced on training night time images. 
    
    \item Ensembling technique has been used to obtain the best performing model. 

    \item The images which contain vehicles rather challenging to detect have been upsampled.
    
    \item Bdd-100k data set has been used to pre-train our model for better identification of cars.
    
\end{enumerate}

\begin{figure*}[t]
    \centering
    \centerline{\includegraphics[width=1.03\textwidth]{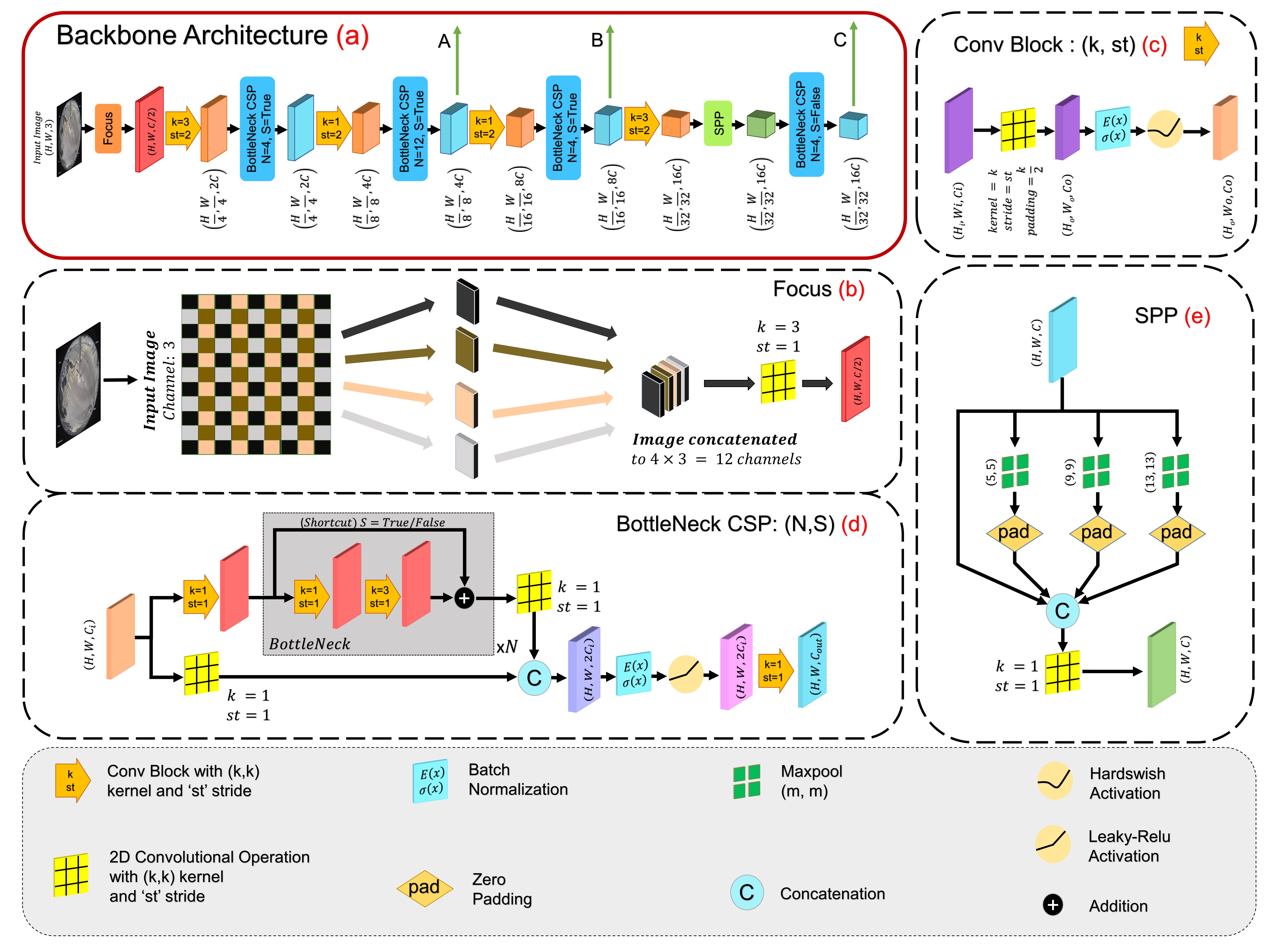}}
    \caption{\textbf{Backbone Architecture of Scheme \textcolor{red}{(a)}. The provided input image is trained through a sequential pipeline consisting of Focus \textcolor{red}{(b)}, Conv Blocks \textcolor{red}{(c)}, BottleNeck CSP \textcolor{red}{(d)} and SPP blocks \textcolor{red}{(e)}; from which three particular feature maps i.e: \textcolor{blue}{A, B, C} are extracted and passed onto the next stage processing which is the Neck.}}
    \label{f1}
\end{figure*}

\section{Proposed Method}

In this particular section, the proposed methodology is presented. For segregating the data distribution, the proposed pipeline introduces a separator model followed by parallel pathways of the core network. After challenging image up-sampling, vehicle generalized transfer learning and vehicle pool increase based data sequence creation, each of the sequences are sent to the pipeline, trained and evaluated. Finally, a selective ensemble ensures the aversion of weaknesses presented by certain data sequences and provides a unified result.

\subsection{Pipeline of the Scheme}
The original dataset is initially partitioned based on day and night as well as upsampled alongside with two additional datasets in order to create various data sequences from them to train on. This constitutes the data preparation part of the scheme shown in Fig~\ref{GA}-A. After obtaining the images from a particular data sequence, they are sent through the network pipeline, shown in Fig~\ref{GA}-B, which is based on two models: The day-night separator and the core model. Images which are meant to be trained are evaluated for being day or night time in order to first create a decision standpoint. Through this, a parallel pathway of two is present and the image in question will move towards one of them. Each pathway consists of the same core model. However, they are trained and optimized for the corresponding time of day. In Fig~\ref{GA}-B, the upper path is optimized for day images and the lower path is optimized for night images. Each data sequence found from the data preparation phase is sent through the pipeline (Fig~\ref{GA}-C) which trains the weights of the model in a different way and produce a different score. These scores are afterwards passed onto the selective ensemble phase of the scheme (Fig~\ref{GA}-D).


\subsection{Implemented Core Network}
%
As discussed before for fast and accurate object detection various YOLO models are considered to be the most suitable \cite{s19030594} \cite{bochkovskiy2020yolov4} \cite{DBLP:journals/corr/abs-1804-02767}.
Recently YOLOv5 has been introduced which is the latest of the YOLO algorithms\cite{repo}.
YOLOv5 introduced a focus block (discussed in details in backbone section \ref{backbone}) instead of the initial convolutional layer present in YOLOv4. The focus block squeezes all the pixels in spatial domain therefore retains more inter-pixel relations and prepares the model for the non-linear distortions introduced at the edge of fisheye images. It is specially effective for vehicle detection in fisheye images since the edge of the image tend to push pixels together. In Table \ref{baseline} the performances of these models on fisheye images are compared. It is evident from this table, YOLOv5 outperforms both YOLOv3 and YOLOv4 while keeping inference time low enough to perform in real time. Therefore, we based our core detection algorithm on the YOLOv5 repository\cite{repo}.

\begin{figure}[h!]
  \centering
  \includegraphics[scale=0.22]{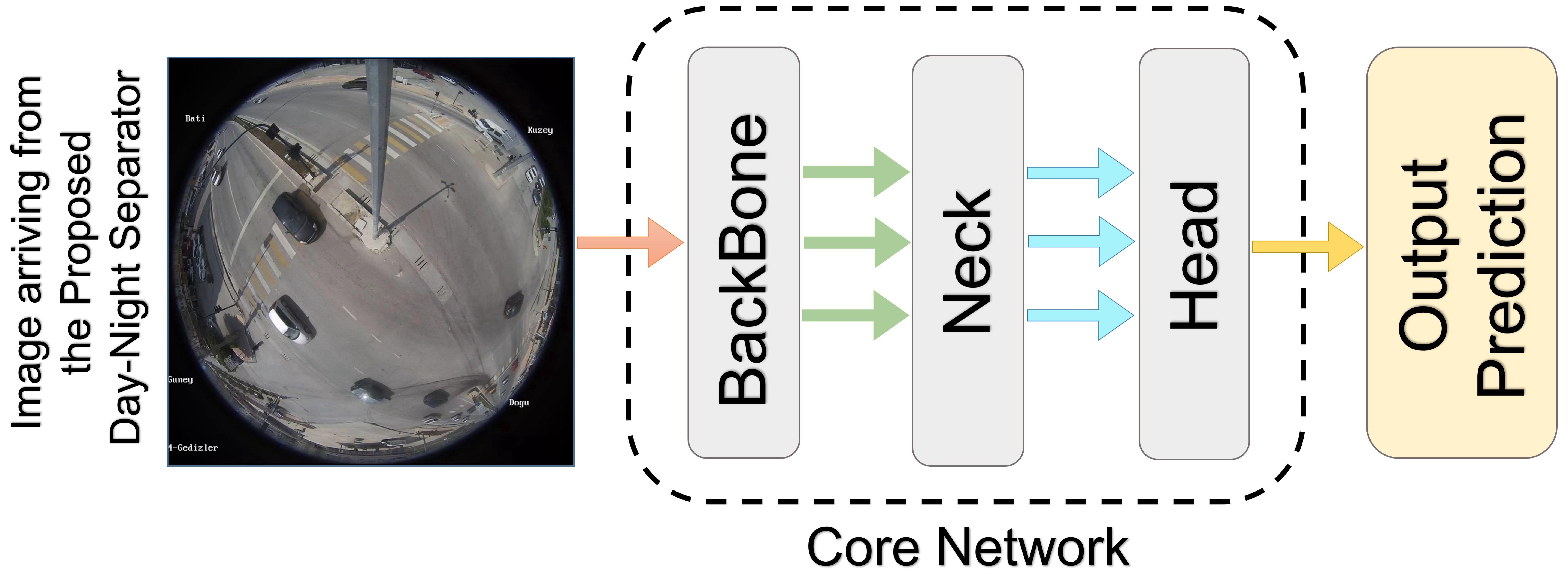}
  \caption{Implemented Core Model}
  \label{core}
\end{figure}

The model comprises of three main parts: backbone, neck and head (Fig. \ref{core}). The main operation occurs at the backbone where the image data are spatially squeezed to form the most significant feature maps. The neck is then used to integrate the low-level features with the high level features extracted from the different stages of the backbone. The head collects the features from the neck and returns the detection results after some adjustments. In what follows each part of the core network is explained in details. 

\subsubsection{Backbone}: \label{backbone}
 The backbone is illustrated in Fig.~\ref{f1}. In order to make the model understand spatial relationships between pixels which are not exactly adjacent to each other, an image is sampled with even (2 in the proposed method) treed gaps on both axes. Each treed gap sampling combination creates a different smaller image and each of them are concatenated channel-wise. This is known as the proposed Focus block (Fig. \ref{f1}) . This is done at the very first part of the backbone architecture. After that, the basic feature learning operation resides in the Conv Block (Fig. \ref{f1}) which also shrinks the feature map spatially and extends it depth-wise. Previous experimentation with Cross Stage Partial Network or, CSPNet architecture \cite{wang2019cspnet} provides intel that splitting the feature vectors into two parts and merging them through a cross-stage hierarchy (conducting simple convolution on one, and numerous Conv Block operation on the other) ensures a strategical gradient flow throughout the network. This operation is performed by the BottleNeck CSP blocks. Furthermore, an additional Spatial Pyramid Pooling or, SPP \cite{He_2014} block is added to increase the receptive field preservation of the network. Three feature maps from the spatially lower side of the backbone are taken (A, B, C) to pass on towards the next stage of the core network. 




\begin{figure}[h!]
  \centering
  \includegraphics[width=\columnwidth]{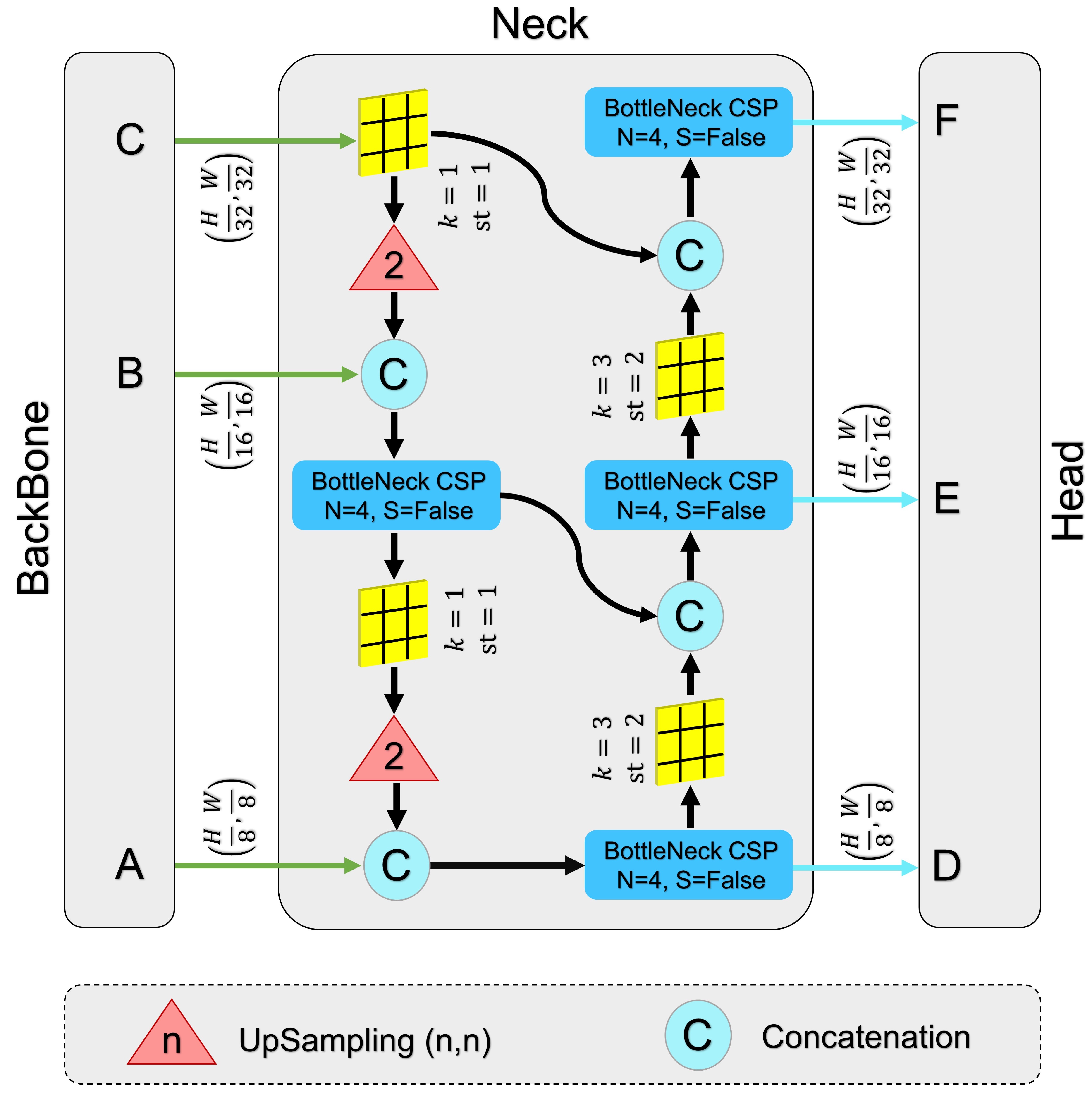}
  \caption{The Neck Architecture: Path Aggregation Network (PANet)}
  \label{panet} 
\end{figure}

\subsubsection{Neck}:
Since vectors arriving from different parts of the backbone possess different shapes, not only the integration of all the higher and lower dimensional features is significant but also ensuring consistency amongst them is necessary. For this, the Path Aggregation Network (PANet) \cite{Liu_2018} is utilized. The lower dimensional vectors are up-sampled spatially and in sequence concatenated with the higher ones. Additionally, a spatial lowering operation is done with re-introduction of the previous up-sampled pathway to create three vectors (D, E, F) which have further correlation among themselves compared to the previous three (A, B, C). The representative network is presented in Fig.~\ref{panet}.

\subsubsection{Head}:
The three feature vectors created at Neck (D,E,F) are passed to this section and further processed to predict the boundary boxes of the vehicles present in the image. For prediction, three anchors are chosen for each of the vectors (D,E,F) by applying K-Means algorithm on training data.  The algorithm applied in YOLOv3  \cite{DBLP:journals/corr/abs-1804-02767} is utilized to calculate the predictions of bounding box parameters (x,y,w,h) and its confidence score using the feature vectors and the anchors. Confidence score of a bounding box specifies the probability of how confident a vehicle prediction is and this is used to threshold and eliminate less sure predictions (confidence threshold). All the predicted boundary boxes are filtered through a Non-max suppression in order to determine the actual boundary boxes.  Then CIoU loss (discussed in a later section) for the predicted boundary boxes and Binary Cross-Entropy loss for confidence and class prediction are finally calculated. 

The YOLOv5 repository contained 4 models: v5s, v5m, v5l and v5x. The models used the same backbone-neck-head scheme detailed so far, but depending on the number of channels and layers used in these models, the parameter count of these models are different. The parameter count of each model is highlighted on Table~\ref{tab:mod-desc}

\begin{table}
\centering
\caption{Model Description}
\label{tab:mod-desc}
\begin{tabular}{|c|c|}
\hline
{\color[HTML]{202020} \textbf{Model}} & {\color[HTML]{202020} \textbf{\begin{tabular}[c]{@{}c@{}}Parameter Count\\ (Million)\end{tabular}}} \\ \hline
{\color[HTML]{202020} YOLOv5s} & {\color[HTML]{202020} 7.3} \\ \hline
{\color[HTML]{202020} YOLOv5m} & {\color[HTML]{202020} 21.4} \\ \hline
{\color[HTML]{202020} YOLOv5l} & {\color[HTML]{202020} 47.0} \\ \hline
YOLOv5x & 87.0 \\ \hline
\end{tabular}
\end{table}

\subsection{Day-Night Separated Training}
\begin{figure}[h!]
  \includegraphics[width=\columnwidth]{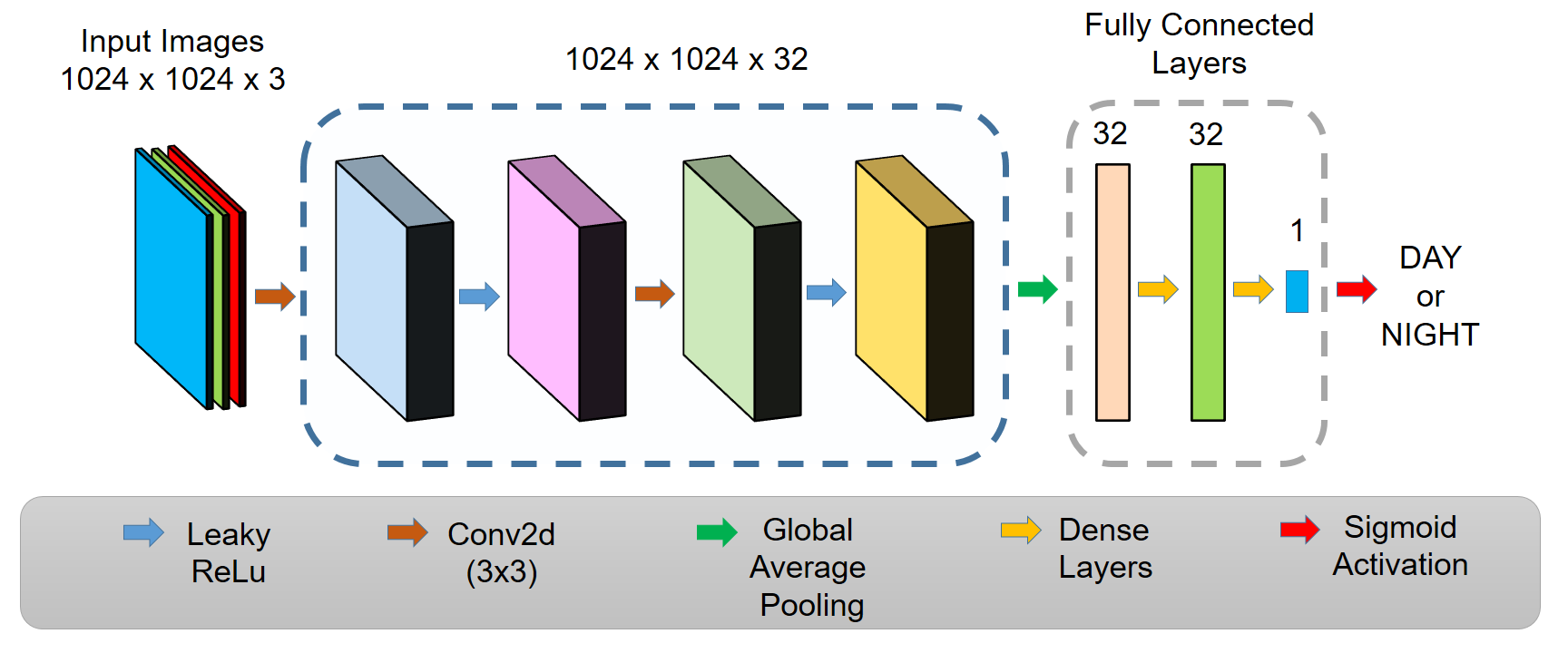}
  \caption{Day-Night Separator Model}
  \label{dvn}
\end{figure}

Due to the lack of traffic activity in the night time compared to the day time, there exists an imbalance between the night time and day time dataset. Moreover, the night time introduces distortions such as exposure from vehicle lights, glare, noise and low brightness. These discrepancies in night and day time datasets lead to night time training affecting the day time training significantly. Moreover, same image appear differently in day and night making it difficult for the same model to generalize for both day and night cases. Thus, training two models, one for day time and one for night time is proposed. To ensure passage of image through the intended night or day model, a day-night separator is proposed which is trained on provided day-night labels. As the day and night images have visible distinguishable features, a shallow convolution neural network as depicted in Fig.~\ref{dvn} is utilized. The separator consists of two convolutional operation with $(3\times3)$ kernels, both having 32 filters. After each convolution, there are LeakyRelu Activations and after the final LeakyRelu, the 3-D features are converted into 1-D fetaures with a global average pooling. Finally, the final day-night classification is done through fully connected layers which provides feedback to the core algorithm whether a day optimized model or a night optimized model is to be chosen for the detection of any particular image. Here it must be noted that in bad weather conditions that is (cloudy or rainy) a day time image would be considered as a night time image. Hence the image would be fed to the night optimized model and the detection will not be affected because detecting exact timestamp of an image is not necessary for our case.

\par

\subsection{Multistage Transfer Learning}
\begin{figure}[h!]
  \includegraphics[width=\columnwidth]{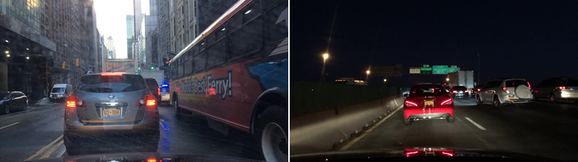}
  \caption{Day and Night Image Instances of BDD100k Dataset}
  \label{bdd}
\end{figure}

The core model is required to distinguish and identify vehicles from a pool of different distributions of not only vehicles but also all the features introduced by a busy city intersection. Therefore, training the network to make it recognize the elemental feature maps of the required vehicles is a necessity and that is where transfer learning comes into action. This process would essentially nudge the local minima of all the model weights towards the global minima of interest provided that the features and contents of each training have somewhat to sufficient similarities with the original data. For this reason, training on a dataset which has similar or sufficiently close features to the goal is required. Since the goal is to detect vehicles in fisheye images, a model which has weights corresponding to object detection or vehicle detection or fisheye images is an initial step for transfer learning. A noted dataset used for this purpose is therefore the COCO dataset~\cite{lin2014microsoft}. The implemented model has a publicly available weight file trained on this dataset which can be directly loaded to the model. The coco dataset however contains different classes including not only bus, car, truck, cycle, bike (of interest), but also person, traffic light, cone, post box, vase, clock (found in a road intersection but are not eligible for detection). Additionally, the coco dataset scarcely include images with vehicle density in roads or intersections. As such, to further make the weights move towards identifying vehicles in roads, an intermediate dataset is utilized, namely BDD100k~\cite{yu2018bdd100k} which consists of images taken from dashboard cameras of moving vehicles. The vehicle based bounding box annotations are taken under consideration and used for training the coco weights loaded model. This ensures the model's putting relevancy on only vehicle features and nothing else before moving on towards the original Fisheye image data training. The multiple weight shifts for the model (from coco to BDD100k to Fisheye) with necessary strides in order to learn the features which are a prerequisite for exact detection makes this strategy a multi-stage transfer learning. A point to note is that BDD100k contains both day and night time images whereas COCO contains very few night cases. So during the training phase, taking which of the day or night time of images or even both based on final prediction is a significant step. In Fig.~\ref{bdd}, an exploratory illustration for the BDD100k dataset is shown from both day and night time.

\par

\subsection{Pseudo Dataset Incorporation}

Although the BDD100k dataset generalizes vehicle features for the core network, it does not generalize fisheye related spherical distortion and the challenges it introduces. In addition, the original data from the VIP Cup 2020 organizer ISSD which is used do not have every type of conceivable vehicles, making detection harder for certain vehicles. In order to further the training procedure from the BDD100k as well as introduce an higher pool of vehicle distribution, a fisheye camera based vehicle dataset of different Thailand road intersections is used \cite{panasonicsecurity_2015} \cite{panasonicsecurity1_2015} \cite{panasonicsecurity2_2015} \cite{panasonicsecurity3_2015} \cite{PanasonicBusiness_2015}. The dataset consists of five videos which are shot with Panasonic 9MP 360-degree security camera (WV-SFV481) at a height of 8m.  Every clip contains video feed from daytime traffic. The video is recorded @ 15 fps with resolution of 3840x2160. For training  all 520 frames are extracted from these videos and resized to 1024x1024 resolution. Since pseudo data can both enrich the dataset and improve the model's performance~\cite{xie2020self}, pseudo labels generated by the model are used and then fed as pseudo dataset for the core model. Since this dataset have similar properties with regards to the original dataset except for the pseudo annotations, it is added with the original data during the training phase.

\begin{figure}[h!]
  \includegraphics[width=\columnwidth]{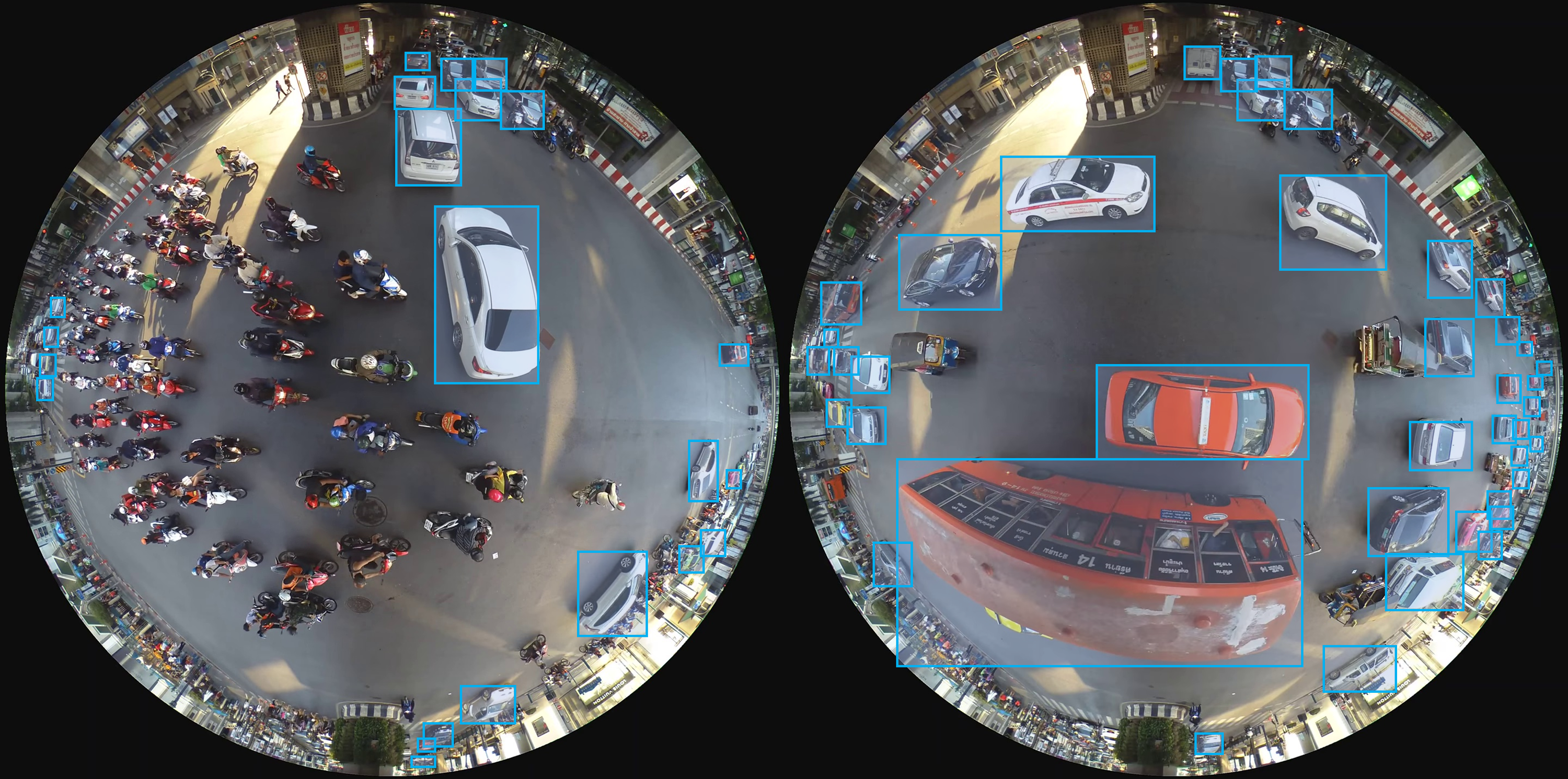}
  \caption{Preview of Pseudo Dataset}
  \label{pseudo-dataset}
\end{figure}

\subsection{Challenging Image Up-sampling}
Imbalanced classes are a common problem in machine learning classification where there are a disproportionate ratio of observations in each class. Most machine learning algorithms work best when the number of samples in each class are almost equal  because most algorithms are designed to maximize accuracy and reduce error. In a fisheye image a vehicle can have different sizes and shapes in different locations of the image. Sometimes a vehicle can have unusual shape because of the fisheye distortion. These cases make the detection task difficult. It is also observed that some objects, for instance cameras mounted into street lamps at junctions are occasionally detected as vehicle due to their shape. As a result, it is needed to make sure that the model learns this type of unusual cases much better. For detecting the challenging cases first images are ranked based on their mAP scores and then a threshold of mAP is set by which the challenging cases are selected. Then the challenging cases are upsampled by 10 times which ensures that the model would face those difficult cases in every epoch. This ensures that the model would learn to recognize these objects better when encountered in the test images.

\subsection{Selectively Ensemble Weights}
Individually speaking, each dataset including the original one contributes unique gradients to the network scheme during training phase. However, the goal of the proposed method is to solely identify vehicles irrespective of environment or day-time, and that is a significant challenge. For this reason, a data partitioning method has been adopted to sequentially train the network keeping in mind the relative surroundings of the present vehicles and image characteristics. The partitions of data are further described in the experimentation section. If an image is of day-time origin, then the network is trained on those subset of datasets in sequence which introduce attributes having vehicles, day time and fisheye curvature; in that particular order. The same argument is applied for the night time images. As a result, different trainings result in different weights for the network. From those weights, the ones which provide the best results with respect to the validation data are selectively chosen and ensembled to get a final aggregated score. Each weights create bounding boxes as predictions through the model and for two different models, overlapped bounding boxes are merged into one but non-intersecting boxes are all included. This would effectively decrease the false-negative prediction and provide enhanced results.  



\subsection{Evaluation Metric}
\textcolor{red}
In order to assess the proposed model’s capability of vehicle detection and localization, mean Precision mean Recall and mAP (mean average precision)  has been used as evaluation matrices. These matrices reflect the precise localization and accuracy of bounding box predictions. 
A true positive case is considered when the predicted bounding box and ground truth bounding box has 0.5 IoU or more. A false positive case consists of predicted bounding boxes having lower than 0.5 IoU with ground truth bounding boxes. 
Change in the confidence threshold brings change in bounding box predictions.  Therefore, different confidence threshold will provide different precision and recall values for the same model.  To summarize these different values, Mean Precision and Mean Recall are used.  Precision refers to the ratio of true positive predictions to the total number of predictions made by the model. Mean precision \cite{everingham2010pascal} is the average of different precision values obtained from changing confidence thresholds. 

Mean Recall \cite{everingham2010pascal} refers to the ratio of true positive predictions and the sum of True positive and False Negative predictions. Mean Recall value is calculated from averaging the recall values obtained for different confidence thresholds. 

The average  value  of  precision  across  all the different recall values, obtained from changing confidence thresholds  is expressed through  mAP(@0.5). \cite{everingham2010pascal}
\begin{align}
mAP = \frac{1}{n}  \sum_{k=1}^{k=n} AP_k 
\end{align}
Here, AP\textsubscript{k} denotes the avarege prevision of class k and n refers to the total number object classes. 

\subsection{Loss}
Three existing loss functions that have been considered for training are GIoU\cite{rezatofighi2019generalized}, CIoU\cite{bochkovskiy2020yolov4} and DIoU\cite{zheng2020distance}. The loss function chosen finally is CIoU. This loss function has a distinct property of factoring into distance and aspect ratio compared to GIoU and DIoU. Generally, the complete loss is defined as follows:
\begin{align}
   CIoU = S(B,B_{gt}) + D(B,B_{gt}) + V(B,B_{gt})
\end{align}

Here, \(B\) and \(B_{gt}\) respectively denote predicted boundary boxes and ground-truth boundary boxes. Furthermore, \(S(.)\) indicate the overlap area similar to GIoU and DIoU and \(D(.)\), \(V(.)\) denote distance and aspect ratio respectively.

CIoU loss acts like a normal DIoU loss when the IoU is less than 0.5; but factors in the aspect ratio when the IoU is bigger than 0.5. This helps CIoU loss perform better in different regression scenarios.

\section{Experimental Setup, Results and Analysis}
In this section, our proposed detection scheme for real time vehicle detection is evaluated on the original VIP Cup 2020 dataset. The proposed detection system is implemented using the deep learning framework Pytorch and run on Google Colab Notebook with an Intel Xeon Processor @ 2.30 GHz, 25 GB of RAM and an NVIDIA P100 16GB GPU.

\subsection{Original VIP Cup 2020 Dataset}
\begin{figure}[h!]
  \includegraphics[width=\columnwidth]{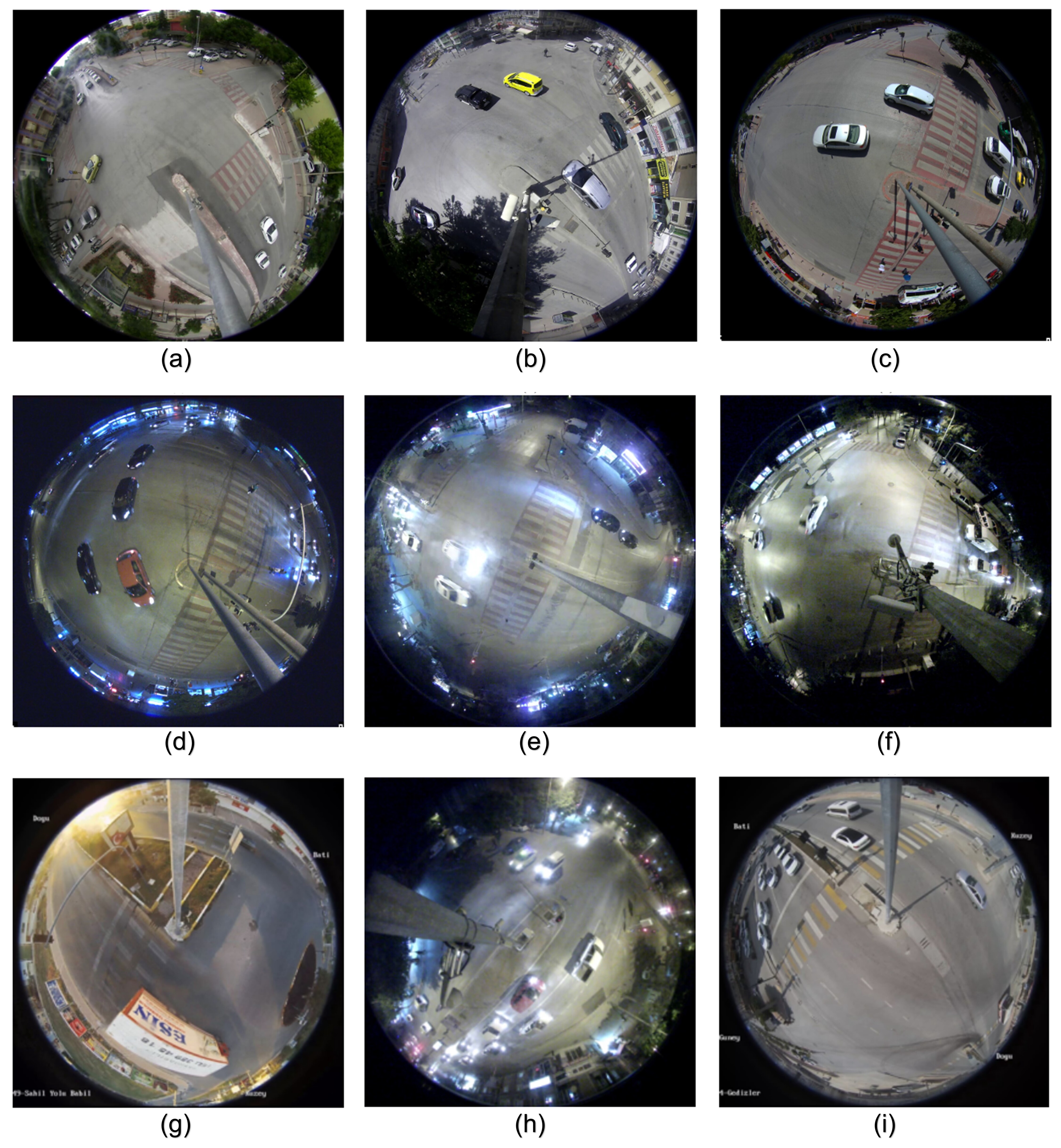}
  \caption{Fisheye Image Instances from the VIP CUP 2020 Fisheye Dataset; which include Day Scenarios \textcolor{red}{(a, b, c)}, Night Scenarios \textcolor{red}{(d, e, f)}, Challenging Scenarios \textcolor{red}{(g, h, i)}.}
  \label{ori_data}
\end{figure}

The original IEEE VIP Cup 2020 dataset contains around twenty five thousand fisheye day images and ten thousand fisheye night images for training and validation purpose and two thousand images for testing. These images are of three different sizes, such as 1024$\times$1024, 1056$\times$1056 and 1280$\times$1280 pixels. The images are collected at different road intersections using fisheye cameras. As a result, the images have non-linear features which make the detection task more difficult. For example, the same vehicle appears at different positions of the subsequent frames with different sizes. The images also encounter lighting variation, shadows, glare and blurry effects. Additionally, there are images which do not possess any vehicles. In Fig~\ref{ori_data}, some examples of the day-night fisheye images from VIP CUP 2020 dataset have been given in which the first three images (a,b,c) are from different junctions during day time and the second three images (d,e,f) show the night views at different junctions. The last three images (g,h,i) imply the difficulties in day and night time situations with fisheye camera. In Fig~\ref{ori_data}(g), high distortion in vehicle structure is observed. The straight shape of the lorry roof has been distorted with a sizable amount of curvature. In Fig~\ref{ori_data}(h), vehicle lights in the night data cause challenging lighting scenarios which can cause distortion of original vehicle features and confuse the object detection model. Different scales of same object moving in different parts of the image complicates the identification of the object across multiple video frames Fig~\ref{ori_data}(i).   

\par

\subsection{Data Partition}
As mentioned before our detection scheme requires training of three datasets. In order to train each of the datasets, they are divided into few partitions. It is important to mention here that the BDD100k contains different types of annotations; namely, object  bounding box, lanes, drivable areas and full-frame instance segmentation. The partitions corresponding to this dataset only includes the bounding boxes with vehicles. The reason behind partitioning each dataset is that although images contained in every dataset have distinct qualities (BDD100k- vehicles, not fisheye; Pseudo- vehicles with fisheye; Original- road intersection vehicles in fisheye), the features and contents in each individual dataset do not follow the same trend throughout. For example, there is lighting discrepancy due to the time period of the day as well as the presence of counter-measures taken by the surrounding area (i.e. headlights, road-lamps, building tube-lights etc).  Therefore, each of them is partitioned in the manner below:

\subsubsection{Fish Day} This partition only contains day time images from the original Fisheye image dataset.
\subsubsection{Fish Night} This partition only contains night time images from the original Fisheye image dataset.
\subsubsection{Fish Mix} This partition contains images of all the day and night time from the original Fisheye image dataset.
\subsubsection{BDD Day} This partition only contains day time images from the BDD100k dataset.
\subsubsection{BDD Night} This partition only contains night time images from the BDD100k dataset.
\subsubsection{BDD Mix} This partition contains images of all the day and night time from the BDD dataset.
\subsubsection{Pseudo} This partition contains images from the incorporated Pseudo dataset which has day time images only.

The image count regarding both the training and validation set for each partition is illustrated using Table~\ref{tab:data-desc}.


\begin{table}[h!]
\centering
\caption{Dataset Description}
\label{tab:data-desc}
\begin{tabular}{|l|c|c|} 
\hline
\diagbox{\textbf{Dataset}}{\textbf{Split}} & \textbf{~ Training ~} & \textbf{~ Validation~~}  \\ 
\hline
\textbf{Fish Day}                          & 20,000                & 1,600                    \\ 
\hline
\textbf{Fish Night}                        & 10,000                & 1,100                    \\ 
\hline
\textbf{Fish Mix}                          & 30,000                & 2,700                    \\ 
\hline
\textbf{BDD Day}                           & 36,000                & 5,000                    \\ 
\hline
\textbf{BDD Night}                         & 27,000                & 4,000                    \\ 
\hline
\textbf{BDD Mix}                           & 63,000                & 9,000                    \\ 
\hline
\textbf{Pseudo}                            & 520                   & -                        \\
\hline
\end{tabular}
\end{table}

\subsection{Day-Night Separator Performance}
\begin{figure}[h!]
  \includegraphics[width=\columnwidth]{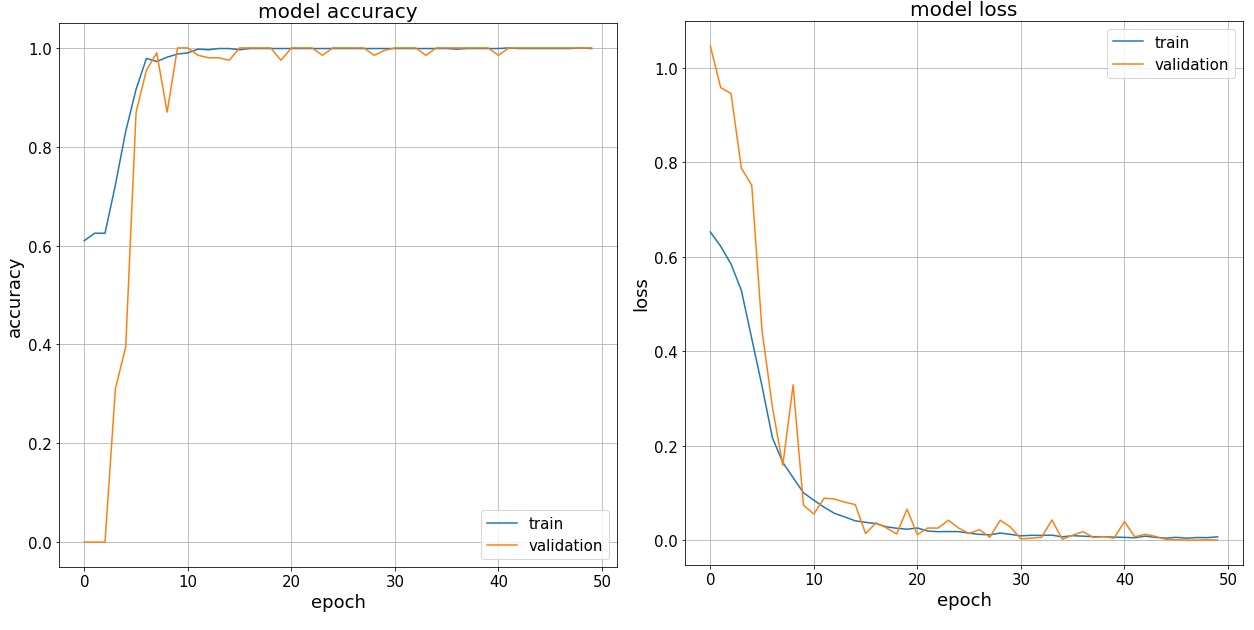}
  \caption{Accuracy and Loss Performances of the Separator}
  \label{sep}
\end{figure}

The day/night classifier needs to be accurate enough to ensure the optimum performance of latter pipelines. As the classifier model is a shallow network, a relatively small sample of the dataset consisting of 500 day and 500 night images is taken for training this model, with a training-validation spilt of 0.8-0.2. The accuracy and loss value for the first 50 epochs are presented in Fig.~\ref{sep}. Afterwards, the trained classifier is evaluated on a test set containing the remainder of day and night cases and an accuracy of 1.0 is achieved here. Therefore, a Day-Night separator is readily obtained for making direct decisions regarding whether an image is day or night.  

\par

\subsection{Core Network Result}
Before performing extensive experimentations on the available datasets, some core models need to be selected to narrow down the number of required experiments. For selecting the core model, various versions of YOLO have been trained and evaluated for images at 416x416 resolution and after that the best ones are selected. Since the chosen models would behave similarly at the same resolution, a less memory intensive approach of selecting 416x416 resolution is utilized. Each of these models have variants with different channels and number of layers while maintaining their distinct architecture. Also, it must be mentioned that the inference time has been considered while selecting the model as it has to perform real time detection for practical usage. The Table~\ref{baseline} indicates the obtained evaluation metrics.    

\begin{table}[h!]
\centering
\caption{Performance chart of Baseline Models}
\label{baseline}
 \begin{tabular}{| c | c | c | c |} 
 \hline
 \textbf{Model}&{\textbf{mAP (@0.5) }}&{\textbf{mAP (@0.5-0.95) }} & \textbf{Inf. Time (ms)} \\  
 \cline{2-3}
 
 \hline
 YOLOv3 & 0.425 & 0.274 & 13.2 \\
 \hline
 YOLOv4 & 0.684 & 0.448 & 10.3 \\
 \hline
 YOLOv5 & \textbf{0.707} & \textbf{0.461} & \textbf{30} \\
 \hline
 \end{tabular}
\end{table}

\begin{table}[h!]
\begin{center}

\caption{{Fish Day Result}}
\label{day}
\begin{tabular}{| c | c | c | c |} 
\hline
\textbf{Resolution} & \multicolumn{2}{c|}{\textbf{Sequential training}} & \textbf{Result of Fish Day} \\  \cline{2-3}
 
 \ & \textbf{a} & \textbf{b} & \textbf{(mAP @0.5} \\  

 \hline
384 & COCO & Fish Day & 0.716 \\
 \hline
512 & COCO & Fish Day &  0.748 \\
 \hline
640 & COCO & Fish Day & 0.797 \\
 \hline
\textbf{768} & \textbf{COCO} & \textbf{Fish Day} & \textbf{0.844} \\
 \hline
 \end{tabular}
 \end{center}
\end{table}


\begin{table*}[t]

 \centering

 \caption{{Network vs Resolution Performance for Original Dataset }}
 \label{5smlx}
 \scalebox{1.0}{
 \begin{tabular}{|c|c|c|c|c|c|c|c|c|c|c|c|}

 \hline
                                   \textbf{Model} & \textbf{Metrics}
                                   & \multicolumn{5}{c|}{\textbf{Image Resolution}}                                                                    \\ \cline{3-7} 
                                   \textbf{Scale}&& \textbf{312}  & \textbf{416} & \textbf{512}  & \textbf{640}    & \textbf{768}     
                                   \\ \hline
 {\multirow{4}{*}{5s}}                            & Inf (sec.)    &     0.0136 &	0.0136 &	0.0140 &	0.0144	& 0.0146 
          \\ 
 \cline{2-7}
                       &   AP          & 0.708 &	0.683 &	0.673 &	0.666 &	0.661 
  \\ 
                       \cline{2-7}
                          & AR          & 0.798 &	0.863 &	0.923 &	0.943 &	0.942 
  \\ 
                          \cline{2-7}
                       & mAP (@ 0.5)          & 0.619 &	0.632 &	0.655 &	0.696 &	0.722 
         \\ 
                       \hline
 {\multirow{4}{*}{5m}}                            & Inf (sec.)         & 0.0168 &	0.0172 &	0.0176 &	0.02 &	0.022	
 \\ 
 \cline{2-7}
                       & AP          & 0.736 &	0.681 &	0.659 &	0.632 &	0.596 
  \\ 
                       \cline{2-7}
                          & AR          & 0.865 &	0.882 &	0.957 &	0.973 &	0.969 
   \\ \cline{2-7}
                       & mAP (@ 0.5)          & 0.621 &	0.64 &	0.663 &	0.708 &	0.745 
   \\ \hline
 {\multirow{4}{*}{5l}}                            & Inf (sec.)         & 0.023 &	0.024 &	0.026 &	0.03 &	0.035 
  \\ 
\cline{2-7}
                       & AP          & 0.744 & 0.723 &	0.715 &	0.694 &	0.681
 \\ 
                       \cline{2-7}
                          & AR          & 0.856 &	0.923 &	0.955 &	0.952  &	0.966 
       \\ \cline{2-7}
                       & mAP  (@ 0.5)          & 0.644 &	0.659 &	0.67 &	0.717 &	0.753
   \\ \hline
 {\multirow{4}{*}{5x}}                            & Inf (sec.)         & 0.027 &	0.03 &	0.033 &	0.045 &	0.053 
  \\ 
 \cline{2-7}
                       & AP          & 0.752 & 0.743 &	0.738 &	0.710 &	0.695 
   \\ 
                       \cline{2-7}
                         & AR          & 0.831  &	0.923 &	0.946 &	0.947 &	0.953 
          \\ \cline{2-7}
                       & mAP  (@ 0.5)         & 0.654 &	0.669 &	0.677 &	0.739 &	0.762 
   \\ \hline

 \end{tabular}
 }
 \end{table*}

\begin{table*}
\centering
\caption{{Fish Night Result (mAP @0.5)}}
\label{night_table}
\begin{tabular}{|c|c|c|c|c|c|c|c|c|c|} 
\hline
\multirow{2}{*}{ \textbf{Reso.} } &\multirow{2}{*}{ \textbf{Label} }& \multicolumn{4}{c|}{\textbf{Sequential Training} }                       &  \multirow{2}{*}{\textbf{Final Stage Result} }  \\ 
\cline{3-6}
                             &     & \textbf{Pretrain}  & \textbf{(i)}  & \textbf{(ii)}   & \textbf{(iii)}     &                                                \\ 
\hline
\multirow{4}{*}{512}           & 512a   & COCO               & BDD Day       & Fish Day        & Fish Night                         & 0.753                                 \\ 
\cline{2-7}
                             &  512b   & COCO               & BDD Day       & Fish Day+Pseudo & Fish Night         & 0.748                                 \\  
\cline{2-7}
                             &  512c   & COCO               & BDD Mix       & -               & Fish Mix+Pseudo    & 0.731                               \\ 
\cline{2-7}
                            &  512d     & COCO               & -             & -               & Fish Night         & 0.518                                \\ 
\hline
\multirow{4}{*}{704}        &  704a     & COCO               & BDD Day       & Fish Day        & Fish Night                     & 0.756                                 \\ 
\cline{2-7}
                            & 704b     & COCO               & BDD Day       & Fish Day+Pseudo & Fish Night       & 0.751                                 \\ 
\cline{2-7}
                            &  704c    & COCO               & BDD Mix       & -               & Fish Mix+Pseudo    & 0.745                                 \\ 
\cline{2-7}
                           &  704d     & COCO               & -             & -               & Fish Night         & 0.546                                 \\ 
\hline
\multirow{4}{*}{768}        & \textbf{768a}     & \textbf{COCO}               & \textbf{BDD Day}       & \textbf{Fish Day}        & \textbf{Fish Night}                      & \textbf{0.760}                                 \\ 
\cline{2-7}
                           &  768b     & COCO               & BDD Day       & Fish Day+Pseudo & Fish Night         & 0.756                                 \\ 
\cline{2-7}
                           &  768c     & COCO               & BDD Mix       & -               & Fish Mix+Pseudo   & 0.748                                \\ 
\cline{2-7}
                            & 768d     & COCO               & -             & -               & Fish Night         & 0.573                               \\
\hline
\end{tabular}
\end{table*}

%

\begin{table*}[]
\centering
\caption{Night Ensemble Result (mAP @0.5)}
\label{night_ensemble}
\begin{tabular}{|l|r|l|r|}
\hline
{\color[HTML]{202020} \textbf{Model Ensemble}} & \multicolumn{1}{l|}{{\color[HTML]{202020} \textbf{Result}}} & {\color[HTML]{202020} \textbf{Model Ensemble}} & \multicolumn{1}{l|}{{\color[HTML]{202020} \textbf{Result}}} \\ \hline
{\color[HTML]{202020} 512a+512b} & {\color[HTML]{202020} 0.756} & {\color[HTML]{202020} 768a+768b} & {\color[HTML]{202020} 0.765} \\ \hline
{\color[HTML]{202020} 512b+512c} & {\color[HTML]{202020} 0.749} & {\color[HTML]{202020} 768b+768c} & {\color[HTML]{202020} 0.758} \\ \hline
{\color[HTML]{202020} 512a+512c} & {\color[HTML]{202020} 0.753} & {\color[HTML]{202020} 768a+768c} & {\color[HTML]{202020} 0.761} \\ \hline
{\color[HTML]{202020} 512a+512b+512c} & {\color[HTML]{202020} 0.757} & {\color[HTML]{202020} \textbf{768a+768b+768c}} & {\color[HTML]{202020} \textbf{0.767}} \\ \hline
{\color[HTML]{202020} 704a+704b} & {\color[HTML]{202020} 0.758} & {\color[HTML]{202020} 768a+704a} & {\color[HTML]{202020} 0.764} \\ \hline
{\color[HTML]{202020} 704b+704c} & {\color[HTML]{202020} 0.751} & {\color[HTML]{202020} 512a+704a+768a} & {\color[HTML]{202020} 0.764} \\ \hline
{\color[HTML]{202020} 704a+704c} & {\color[HTML]{202020} 0.756} & {\color[HTML]{202020} 768a+768b+768c+704a} & {\color[HTML]{202020} 0.766} \\ \hline
{\color[HTML]{202020} 704a+704b+704c} & {\color[HTML]{202020} 0.758} & {\color[HTML]{202020} 768a+768b+768c+704b} & {\color[HTML]{202020} 0.766} \\ \hline
\end{tabular}
\end{table*}

\subsection{Network and Resolution Oriented Result}
To establish a baseline of the YOLOv5 models referenced at Table~\ref{tab:mod-desc}, the models are trained on the fish-mix data partition (consisting of all the day and night images from the original Fisheye dataset). In the YOLOv5 models, the height and width of the output feature map(generated from image input) gets reduced by a total factor of 32 by the downsampling layers. So the models are trained on different image resolutions that are multiples of 32 and the effect of changing image resolutions in each model is also examined. The results of the experiments are detailed in Table~\ref{5smlx}. It is observed from the table that as the resolution increases for a particular model variant, the average precision (AP) decreases but the average recall (AR) increases. This proves that for higher resolution, the model detects higher number of False-Positive and low quantities of False-Negative. The reason behind this is that the provided original data had some mislabelled ground truth. After visual analysis, it is found that a few vehicles are unlabelled in some images. The trained model misses fewer vehicles with the increase of the image resolution. This can be attributed to the larger and more detailed feature map generated by the model. As a result, the metrics treat many of the model's detection as false-positives where the vehicle is unlabelled in ground truth. Furthermore, for low resolutions, the model misses vehicles which are far away, blurry or under shadows. And these missed detection coalesce with some of the aforementioned unlabelled vehicles in ground truth and prevents false-positives to occur in small resolution images.


\subsection{Training Result of Day and Night data}


To evaluate the models' performance on the day and night case, they have been trained on different combinations of the datasets and evaluated on the Fish-Day and Fish-Night data. For day cases initially a baseline YOLOv5 model is selected which is trained on Fish-day data only, as shown in Table~\ref{baseline}. After selecting the best model, first the model is initialized with COCO pre-trained weights and then trained on Fish-Day data. Performances of the proposed model on different combinations of the Fish-day data are provided in Table~\ref{day}. 

Similarly for the night cases the models have been initialized using COCO dataset weights to take advantage of the transfer learning. Performances of the proposed model on different combinations of the Fish-Night data is shown in Table~\ref{night_table}. The network has consistently displayed improved performance as the training image resolution is increased. Each sequence of training is labelled in the second column of the table (512a, 512b, 704a etc) for later reference. Inclusion of both BDD and Pseudo dataset results in around 1\% and 0.1\% improvement in the day score respectively. But it does not improve the score for night cases as Pseudo data has only day images, which creates more bias towards the day cases. As there are little data in Fish-Night compared to Fish-Day, model trained on Fish-Day can extract some unique features that Fish-Night model can not extract. Thus model pre-trained on Fish-Day and then trained on Fish-Night yields the best results. The model which has been pre-trained on both Fish Day and Night (Fish-Mix) performs little less than the model that has been trained on the only Fish-Day because model struggles to learn both day and night features simultaneously as they are very different. Furthermore, the model that has been trained on only Fish-Night after COCO, does not perform as good as the others as there are not enough samples for Fish-Night cases compared to the Fish-Day cases.

\begin{figure*}[h!]
\centering
  \includegraphics[width=0.90\textwidth]{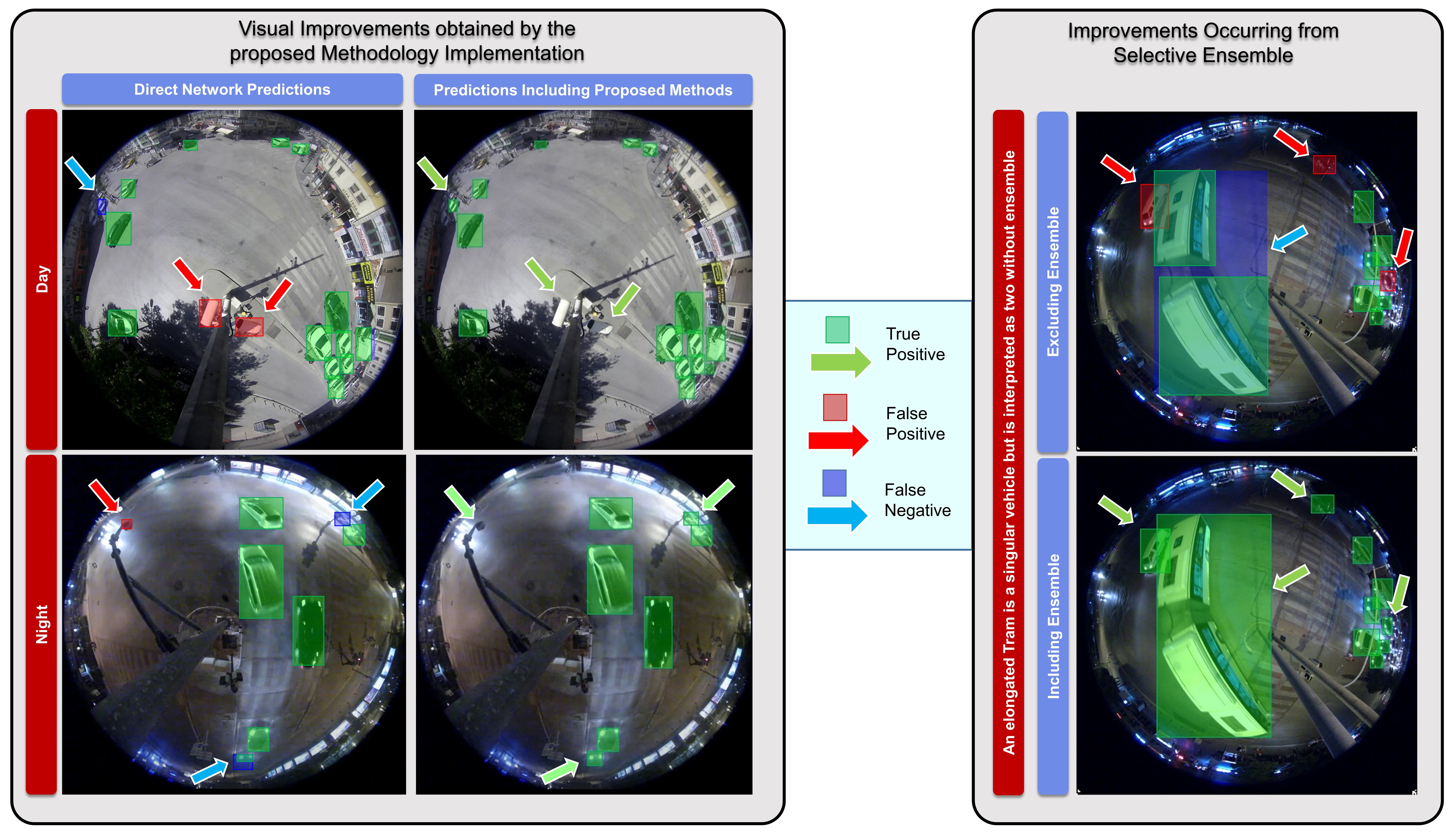}
  \caption{Illustration of improved predictions obtained from including proposed schemes. The introduced training methods reduce false detection and missed detection of vehicles; whereas the ensemble ensures the identification of complicated vehicles.}
  \label{improv}
\end{figure*}

\subsection{Selective Ensemble Result}
Observing the performance of models on the Fish-Day and Fish-Night data (from Table~\ref{day} and Table~\ref{night_table} respectively), it becomes evident that improvement of the result in night cases is very challenging. In previous section, schemes a,b,c and d are introduced as shown in Label column in Table~\ref{night_table} based on the combinations of pretraining on COCO, BDD-Day or Mix, Fish-Day and Pseudo dataset. Then finally the Fish-Night data is fed into the pre-trained model to acheive better results. To further improve the performance in night cases, the models trained on these schemes (detailed on Table~\ref{night_table}) are ensembled and the results are observed in Table~\ref{night_ensemble}. Only the combinations of scheme a, b and c are shown, since scheme d performs poorly during ensemble. 

From the Table~\ref{night_ensemble}, it can be seen that ensembling scheme a and b provided around 0.3-0.5\% improvement over the single models. Ensembling with any model trained on scheme c contributed little to the final result, which can be attributed to the individual low performance resulting from scheme c. The highest result is achieved by using an ensemble of three models trained on resolution of $(768\times768)$. Ensembling more models with the highest scoring ensemble does not increase the results further, which implies that the models trained on schemes a, b and c already has the information that the new models could provide. 

\subsection{Visible Procedural Improvements}
In Fig.~\ref{improv}, the visual improvements are shown based on the implementation of the proposed scheme as well as excluding the scheme entirely. The direct predictions from the network results in false-positives of traffic lights in contrast to vehicles and false-negatives of visible vehicles which are far away or partially behind other vehicles. Day-Night separation, up-sampling, multistage transfer learning and pseudo data incorporation altogether decreases the existing false-positives and false-negatives. Furthermore, the selective ensemble process can detect complex vehicle structures, shown in Fig.~\ref{improv} (a tram), which the network alone detects incorrectly (detecting it as two individual vehicles) without the introduced methods.

\section{The Ground Truth Inconsistency Issue in the Original Dataset}

\begin{figure}[h!]
\centering
  \includegraphics[width=0.4\textwidth]{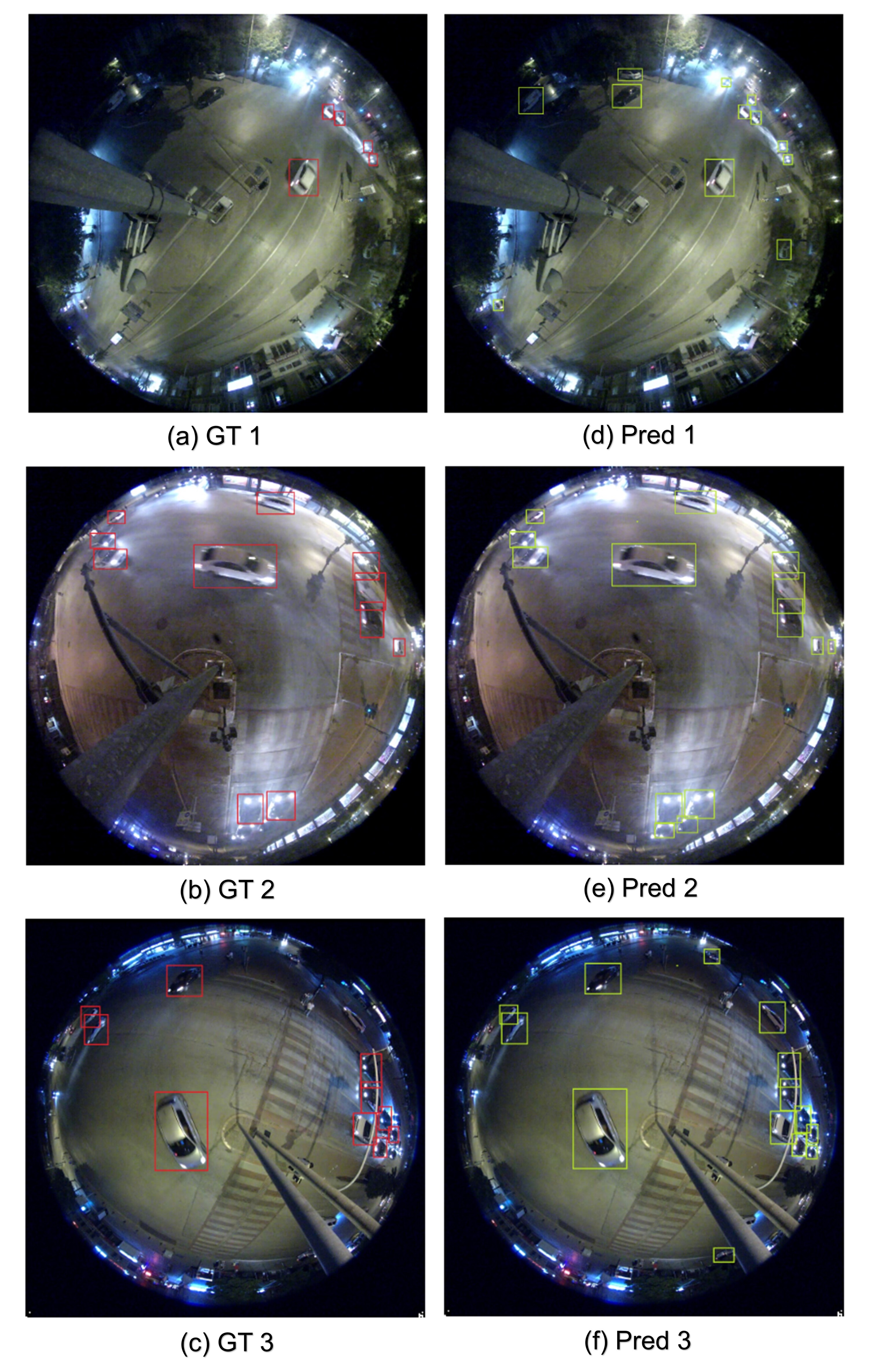}
  \caption{Three instances of ground truth bounding-boxes \textcolor{red}{(a, b, c)} provided in the Dataset. The proposed methodology provides detection \textcolor{red}{(d, e, f)} of the corresponding images which are visually more accurate compared to the ground truths.}
  \label{incon}
\end{figure}

The original dataset has some inconsistent ground truths e.g. some vehicles near the image circumference, under shadows or subject to flares are not annotated. These noisy labels cause a rise in false positives which attributes to a lower evaluation score because the average precision (AP) drops noticeably. Additionally, the core model generalizes vehicles very decently which in turn results in very little false negative occurs. This issue is more problematic in higher resolutions where the network misses even less vehicles even if they are far away at the edges of the fisheye images. Such ground truth inconsistency resides in mostly night time images rather than day time images. The result showcased in Table~\ref{5smlx} agrees with this incident (outcome for day-night mixed images) and it is observed that the AP values are low compared to the AR values. A few noisy labels and corresponding predictions are shown in Fig.~\ref{incon}. The ground truth bounding-boxes marked in \textcolor{red}{red} have missed some vehicles but the model's prediction marked in \textcolor{green}{green} did not miss them.

\par

\section{Conclusion}
In this paper, a YOLOv5 based vehicle detection scheme has been proposed for fisheye images under both day and night conditions. Various object detection algorithms have been tested and YOLOv5 turned out to be the best suited for real-time vehicle detection in fisheye images. Introduction of the day-night differentiator network enabled us to increase the accuracy of vehicle detection in both day and night time images without compromising speed. The ensembling method ensures that the  best trained weights are used for inference. Vehicle detection on the fisheye images achieved good accuracy and adequate speed. Extensive experimentation on the original VIP Cup 2020 dataset shows our method outperforms even YOLOv5 algorithm on accuracy and achieves real-time detection.




\bibliographystyle{IEEEtran}
\bibliography{ref}
\end{document}